\documentclass[runningheads]{llncs}
\usepackage[
backend=biber,
style=numeric,
sorting=none
]{biblatex}

\usepackage{hyperref}
\usepackage{algorithm}
\usepackage{algpseudocode}
\usepackage{graphicx}
\usepackage{setspace} 
\usepackage{amssymb}
\usepackage[misc,geometry]{ifsym}

\begin{document}

\title{Noise transfer for unsupervised domain adaptation of retinal OCT images}

\author{Valentin Koch\inst{1,3,5} \and
Olle Holmberg \inst{1,2} \and Hannah Spitzer\inst{2} \and Johannes Schiefelbein\inst{4}\and Ben Asani\inst{4} \and Michael Hafner\inst{4} \and Fabian J Theis\inst{1,2} \Letter}

\authorrunning{V Koch et al.}

\institute{Technical University of Munich, Munich, Germany
\email{fabian.theis@helmholtz-muenchen.de}
\and Institute of Computational Biology, Helmholtz Munich, Munich, Germany \and
Institute of AI for Health, Helmholtz Munich, Munich, Germany \and 
Department of Ophthalmology, Ludwig-Maximilians-University, Munich, Germany \and
Munich School for Data Science, Munich, Germany
}
\maketitle 
\begin{abstract}
Optical coherence tomography (OCT) imaging from different camera devices causes challenging domain shifts and can cause a severe drop in accuracy for machine learning models. In this work, we introduce a minimal noise adaptation method based on a  singular value decomposition  (SVDNA) to overcome the domain gap between target domains from three different device manufacturers in retinal OCT imaging. Our method utilizes the difference in noise structure to successfully bridge the domain gap between different OCT devices and transfer the style from unlabeled target domain images to source images for which manual annotations are available.  We demonstrate how this method, despite its simplicity, compares or even outperforms state-of-the-art unsupervised domain adaptation methods for semantic segmentation on a public OCT dataset. SVDNA can be integrated with just a few lines of code into the augmentation pipeline of any network which is in contrast to many state-of-the-art domain adaptation methods which often need to change the underlying model architecture or train a separate style transfer model. The full code implementation for SVDNA will be made available at \url{https://github.com/ValentinKoch/SVDNA}.

\end{abstract}
\section{Introduction}

Diseases in the Human retina are among the leading reasons for reduced vision and blindness globally. Estimates are that currently roughly 170 million people are affected by Age-related Macular Degeneration \cite{Pennington2016-vj}, while Diabetic Retinopathy is recognized as a global epidemic with the numbers increasing at ever higher rates \cite{Lee2015-eu}. Optical coherence tomography (OCT) is a powerful technique used in many medical applications to generate real time and non-invasive cross-sectional images of live biological tissues \cite{Al-Mujaini2013-ww}. In the field of Ophthalmology, OCT images help doctors to make therapy decisions and monitor the treatment outcome. As eye diseases are becoming more and more prevalent due to the increased age of populations \cite{Fricke2018-ho},  the need for research in automating diagnosis and aiding doctors is increasing.

Recently, deep learning methods are showing promising results in areas such as disease prediction \cite{Banerjee2020-zd, Holmberg2020-rs}, semantic segmentation \cite{Borkovkina2020-cq, Hassan2021-zf} or improving quality \cite{Cheong2021-qc} of retinal OCT images.  Although a lot of progress has been made, challenges remain in applying artificial intelligence methods on real-world OCT data, where image characteristics such as signal-to-noise ratio, brightness, and contrast can vary and cause changes in data distribution, so-called domain shifts. For OCT images, a particular domain shift is introduced from different OCT imaging devices being used, which have different image-quality properties. While for a medical doctor those differences are only a mild annoyance, machine learning models can quickly fail when faced with only small disturbances in the underlying data distribution. One solution is to label images from all possible devices, but as manually labeling images is very costly and needs highly skilled specialists, other methods need to be developed.

Domain adaptation methods offer a solution to the reduced performance of AI algorithms that are caused by the difference in data distribution \cite{ Guan2021-wh, Patel2015-bn}. 
Domain adaptation has also been used for shifting domains between different device manufacturers for OCT imaging devices. Yang et al. \cite{Yang2020-cq} detect lesions in OCT images from different camera devices using an adversarial approach, several recent works use CycleGAN approaches to transfer style between domains  \cite{Romo-Bucheli2020-az, Zhang2020-mu, Manakov2019-cw}. In particular, Romo-Bucheli et al. train a CycleGAN and measure the improved performance on a segmentation task, which makes it the most comparable work to ours \cite{Romo-Bucheli2020-az}. While GAN architectures are performant when dealing with domain shifts where the structural difference between the domains is much larger than between OCT camera devices \cite{yang2020phase}, we argue that for domain adaptation between retinal OCT devices these models are unnecessarily complex and can through small changes to image content even be decremental to the performance.

In this work, we present a novel method for unsupervised domain adaptation (UDA) for semantic segmentation of retinal OCT images from multiple devices. The underlying observation motivating our approach is twofold. First, we observe that the noise structure is a key difference between images from different OCT cameras and needs to be specifically modeled. Second, we observe that many UDA methods are often developed for unsupervised domain adaptation between synthetic and real-world data and are not necessarily optimal for retinal OCT images. We therefore develop and evaluate a simple but effective unsupervised domain adaptation method using a singular value decomposition-based noise adaptation approach (SVDNA). We train a semantic segmentation model by using SVDNA and show that the model generalizes to unseen OCT devices and even performs on par with supervised methods trained with manual labels. Further, we show that the SVDNA method is comparable or even outperforms other state-of-the-art UDA methods that often require more complex training schemes or implementation and usage of separate style transfer models. Further, to the best of our knowledge previous work evaluated on private datasets, making comparisons difficult, whereas we benchmark SVDNA on  publicly available OCT datasets from multiple devices.

\subsubsection{Contributions} 
We make the following contributions to OCT imaging analysis and biomedical unsupervised domain adaptation:
\begin{itemize}
    \item We present SVDNA, a minimal method for unsupervised domain adaptation method that transfers style between images by using a singular value decomposition-based noise transfer model.
    \item We demonstrate that our method performs on par or even outperforms more complex state-of-the-art UDA methods as well as models trained with supervised labels, while considerably reducing training complexity.

\end{itemize}

\section{Proposed method}
First we introduce our Singular Value Decomposition-based noise adaptation method. In the second part, we describe how we trained our segmentation network with the help of the SVDNA restyled images.

\begin{figure}
\includegraphics[width=\textwidth]{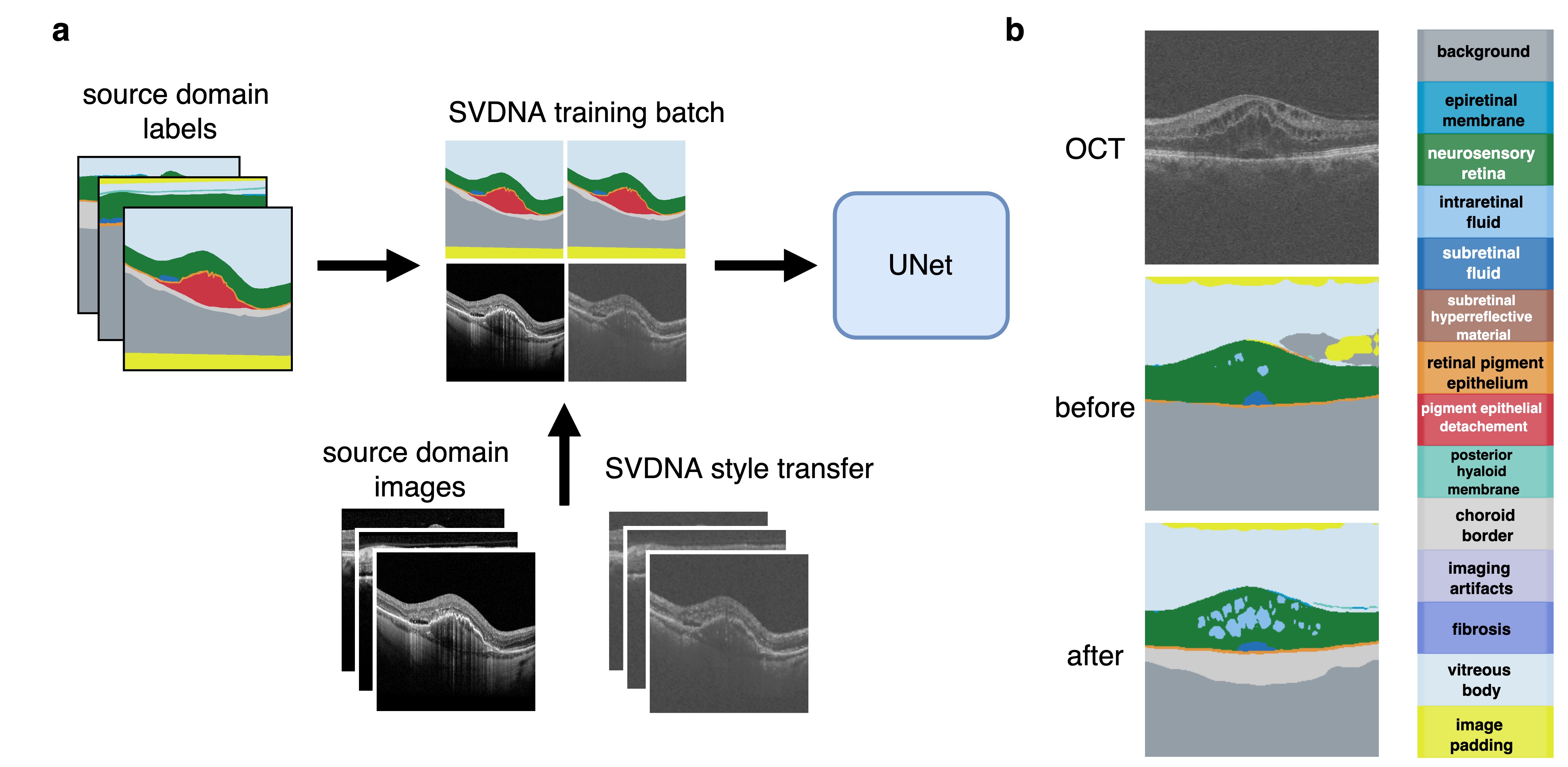}
\caption{ (a):We train a segmentation network with images from the Spectralis device which are randomly restyled to the style of the Cirrus, Topcon or Bioptigen OCT device using SVDNA, thereby enabling unsupervised adaptation to multiple domains. (b)  Comparisons between segmentations before and after SVDNA domain adaptation.
} \label{fig:1}
\end{figure}

\subsection{Singular value decomposition-based noise adaptation (SVDNA)}
SVDNA achieves style transfer between a source and target domain image by decomposing both images of size $n\times n$ into their respective singular value decompositions $U\Sigma V^T$, where $U$ corresponds to the left singular vectors, $V$ to the right singular vectors and $\Sigma$ to the singular values. Then, the reconstructed noise from the target domain image is added to the reconstructed content of the source image. Therefore, we use the first $k$ singular values and their corresponding right and left singular vectors from the source image, where the content is encoded, and add the noise that is encoded in the $k+1,..., n$ singular values and their corresponding vectors from the target image. The resulting image can be expressed as a matrix multiplication as can be seen in algorithm \ref{alg:1}. The parameter $k$ must be chosen by hand but is in our experience not very sensitive to variation. In practice, values between $k=20$ and $k=50$ were used to train our network with images of $256\times 256$ pixels. For more details on the feasibility of the used values of $k$ see supplementary figure 2.
\begin{algorithm}
\caption{SVDNA}\label{alg:cap}
\begin{algorithmic}
\setstretch{1.35}
\State Let $U\Sigma V^T$ be the singular value decomposition of the respective images im$_{source}$, im$_{target} \in \mathbb{R}^{n\times n}$ and $k$ the noise transfer threshold. 
\\\hrulefill
\State im$_{source}=U_s\Sigma_sV_s^T, \quad $im$_{target}=U_t\Sigma_tV_t^T$,
\State $U_r \gets u_s^1, ..., u_s^k,u_t^{k+1}, ..., u_t^n$
\State $\Sigma_r\gets $diag$(\sigma_s^1, ..., \sigma_s^k, \sigma_t^{k+1}, ..., \sigma_t^n) $
\State $V_r^T \gets v_s^{1^T}, ..., v_s^{k^T},v_t^{{k+1}^T}, ..., v_t^{n^T}$  
\State Im$_{noised} \gets U_r\Sigma_rV_r^T$  
\State Im$_{clipped} \gets $clip\_values\_to\_interval$($Im$_{noised}$,$[0,255])$
\State Im$_{restyled\_final} \gets $histogram\_matching$($source$=$Im$_{clipped}$, target$=$im$_{target})$

\end{algorithmic}
\label{alg:1}
\end{algorithm}
To account for possible out of bound pixel values that can occur after this noise transfer operation, values are clipped to the interval $[0,255]$. In addition, a histogram matching  \cite{Gonzalez2009-ky} step is done after the noise adaptation to match pixel intensity distribution. This step is motivated by the fact that after the addition of target image noise and source image content, the pixel values of the resulting restyled image and target image are still differently distributed. The effect of this step can be seen in the ablation study in table 1 of the supplementary material.

Combined, this does not only achieve a visually good style transfer, but we are also able to match different noise-related metrics of the target domain well as seen in figure \ref{fig:umap}. We evaluate the noise transfer from our private source domain dataset, where images were taken with a Spectralis device, on three different datasets: Two datasets from the RETOUCH challenge \cite{Bogunovic2019-hh} \footnote{Access can be requested at \url{https://retouch.grand-challenge.org/}}  who use Cirrus and Topcon devices as well as a dataset with images taken with a Bioptigen device \cite{Farsiu2014-xv} \footnote{\url{https://www.kaggle.com/paultimothymooney/farsiu-2014}}.

\subsection{Training the segmentation network with SVDNA}
SVDNA can be used to train any segmentation network and is applied before augmenting the training images. When loading an image, either no style transfer (probability $p= 1/n$), or an SVDNA style transfer to a randomly chosen target domain ($p=1-1/n$) is applied, where $n$ is the total number of domains, including the source domain. When a target domain is chosen, one image is randomly selected and used as a style target for the input image. When the source domain is chosen, no style transfer is applied. To maximize style variability, the hyperparameter $k$, determining the amount of noise to be transferred, is randomly sampled for each style transfer individually within range $[20,50]$. As the content of the source image is combined with the style of the target image, the annotated labels of the source dataset can be used as ground truth to train the network.

\section{Results}
\begin{figure}
\includegraphics[width=\textwidth]{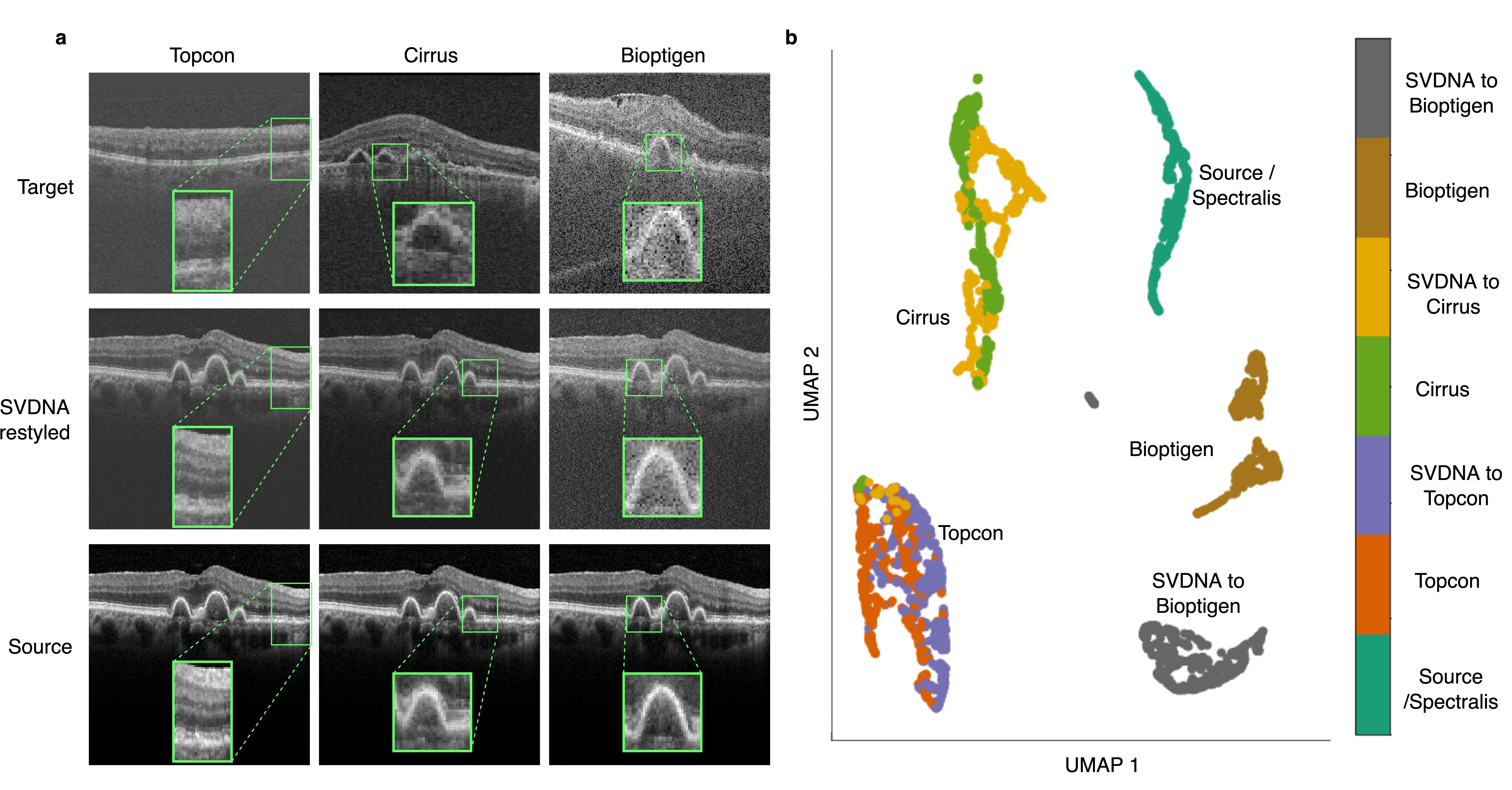}
\caption{(a): Sample SVDNA restyling to Topcon, Cirrus, and the Bioptigen device from a single source domain image (bottom row) for $k=30$. (b): Noise statistics UMAP \cite{mcinnes2020umap} between datasets of different domains and SVDNA adapted datasets.} \label{fig:umap}
\end{figure}
We compare our SVDNA against state-of-the-art domain adaptation approaches that we trained with the same segmentation model architecture, data processing pipeline, and augmentations. As the baseline, we use a network without any domain adaptation, which we compare to Fourier Domain Adaptation (FDA) \cite{Yang2020-vd}, Confidence regularized self-training (CRST) \cite{Yang2019-ik}, the CycleGAN approach \cite{Romo-Bucheli2020-az} and an SVDNA trained model.  For each method, the main hyperparameters were individually optimized to include each models best possible results in the comparison. Training details of all methods can be seen in the supplementary material.
\subsection{Data}
Our source domain training set consists of 462 OCT scans of the macula, taken with a Spectralis device (Spectralis; Heidelberg Engineering GmbH, Heidelberg, Germany) from different patients suffering from age-related macular degeneration. It is a private dataset annotated by three retinal experts of the Department of Ophthalmology, Ludwig-Maximilians-University, Munich (B.A., J.S. and M.H.), where each pixel of an OCT scan is labeled with one of 14 classes following the Consensus Nomenclature for Reporting Neovascular Age-Related Macular Degeneration Data of the AAO (American Academy of Ophthalmology) \cite{SPAIDE2020616}: Intraretinal Fluid, Subretinal Fluid, Pigment Epithelium Detachments, Fibrosis, Epiretinal Membrane, Posterior Hyaloid Membrane, Subretinal Hyper Reflective Material, Neurosensory Retina, Choroid layers, Choroid Border, Vitreous and Subhyaloid Space, Retinal Pigment Epithelium, imaging artifacts, image padding. 
For the quantitative evaluation, the RETOUCH challenge dataset is used, where images are taken from the Topcon ($2688$ images) and Cirrus ($3072$ images) device and are annotated with Subretinal Fluid, Intraretinal Fluid, Pigment Epithelium Detachments. The OCT images annotations were not used in training any algorithms but only for evaluating performance.

\subsection{Evaluation}
\begin{figure}
\includegraphics[width=\textwidth]{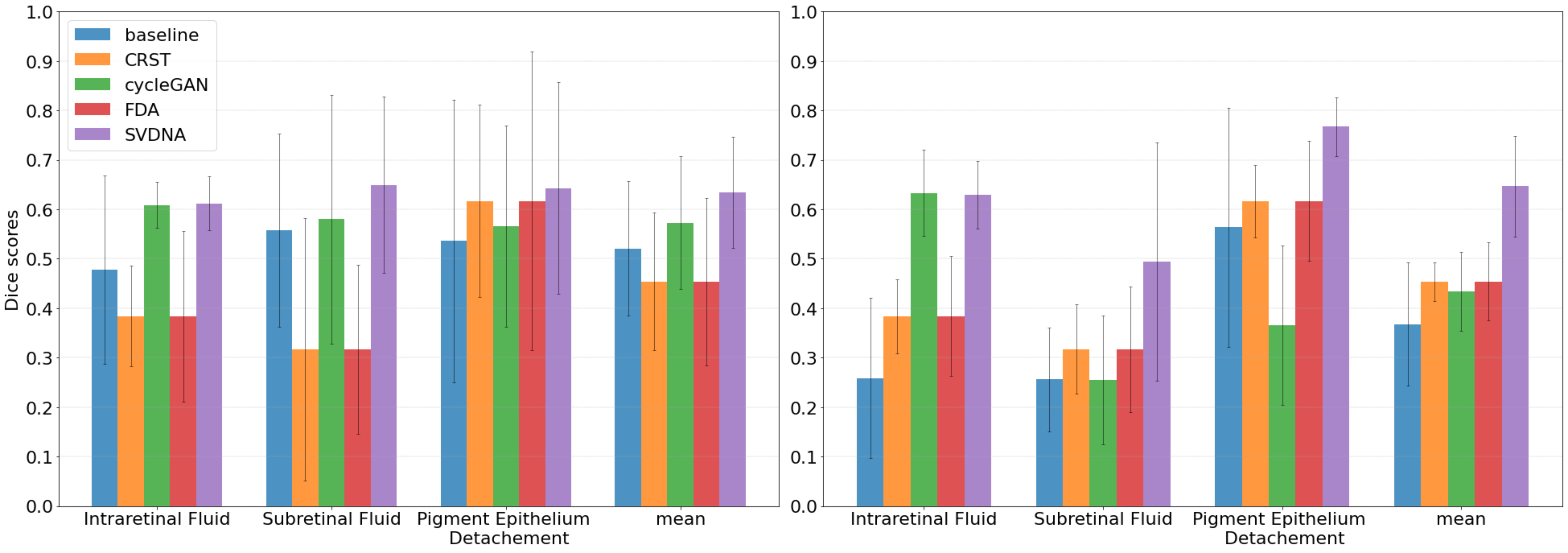}
\caption{5-fold cross-validation comparison of different state-of-the-art methods, the SVDNA method, and the baseline network on the RETOUCH challenge datasets Cirrus (left) and Topcon (right). The SVDNA trained model outperforms the baseline and compares or outperforms all other tried methods across all datasets and classes.
} \label{fig:bar}
\end{figure}
For all experiments a Unet++ \cite{Zhou2018-ab} with a ResNet18 \cite{He2016-ux} encoder is used as the segmentation model. 
In figure \ref{fig:umap}a, SVDNA is applied to images of three different domains, showing how one image can be fitted to the style characteristics of different target domain images. Figure \ref{fig:umap}b shows a noise representation UMAP \cite{mcinnes2020umap} embedding. For the embedding, three different noise statistics (signal-to-noise ratio \cite{Janesick2007-wo}, noise variance estimator \cite{Immerkaer1996-og}, and wavelet noise standard deviation estimator \cite{Donoho1994-gx}) are used. After SVDNA, the noise embeddings of the restyled Spectralis images align closely with those of the target domains Topcon and Cirrus. Only with Bioptigen, a domain with a very high noise level, there is still a gap between the respective embeddings. 
Additionally, we compare SVDNA against state-of-the-art domain adaptation approaches for the task of semantic segmentation on two datasets consisting of images of a Topcon or Cirrus device respectively in figure \ref{fig:bar}. As the baseline, a network without any domain adaptation is used, which is compared to Fourier Domain Adaptation (FDA) \cite{Yang2020-vd}, Confidence regularized self-training (CRST) \cite{Yang2019-ik}, the CycleGAN approach of Romo-Bucheli et al. \cite{Romo-Bucheli2020-az} as well as  our SVDNA trained model. For each method, the individual hyperparameter were optimized to include each models best possible results in the comparison. For further training details of all methods, we refer to the supplementary material. The evaluation is done on a 5-fold cross-validation over both datasets, using 80\% of the target domain images as style targets for SVDNA or the Fourier Domain Adaptation or as training images for Self-Training and evaluate on the 20\% remaining images and iterate this until the algorithm has been evaluated on all data samples. The CycleGAN method was, due to its complicated proposed evaluation scheme, trained on all of the images. As done in the RETOUCH challenge, we measure the performance with the dice score.
The largest performance gain over the baseline can be seen for images from the Topcon device, where the SVDNA model consistently outperformed all other methods segmenting subretinal fluid and PED and is on par with the CycleGAN method for intraretinal fluid. When considering the mean performance difference across all classes, the SVDNA model performs better than all other methods.

\begin{table}
\caption{SVDNA and baseline model compared to the best supervised trained models from 8 teams in the RETOUCH challenge \cite{Bogunovic2019-hh} on a non-public testset of Topcon (left) and Cirrus (right).}\label{tab1}
\begin{tabular}{llllll}
Name &  SRF & IRF & PED & mean & rank\\
\hline
SFU \cite{LU2019100} &  \underline {0.80} & \underline  {0.72} & \underline {0.74} & \underline {0.75} & \underline 1\\
 SVDNA &  \underline {0.80} &0.61 &0.72 & 0.710 & 2\\
 Baseline &  0.42&  0.17&  0.66& 0.42& 10 \\
\hline
\end{tabular}
\quad
\begin{tabular}{llllll}
Name &  SRF & IRF & PED & mean & rank\\
\hline
SFU \cite{LU2019100} &  \underline {0.72} & \underline {0.83} &  0.73 & \underline {0.76} & \underline 1\\
 SVDNA &  0.66 &  0.61 &  0.74 &  0.67 &  7\\
 Baseline &  0.42&   0.39& 0.71&  0.51 &  10 \\
\hline
\end{tabular}
\end{table}

We also benchmark SVDNA on the separate hold-off non-public test datasets, where we compare our domain adaptation method to results achieved by fully supervised trained networks submitted in the RETOUCH challenge. We handed in predictions of the naive baseline as well as from our SVDNA trained model, again on the three classes Subretinal Fluid (SRF), Intraretinal Fluid (IRF) and Pigment Epithelium Detachements (PED). With SVDNA we achieved the second-best result of 10 submitted segmentation methods on the Topcon dataset and the sixth-best on the Cirrus dataset. SVDNA achieved a large improvement over the baseline, as well as a very competitive performance compared to fully supervised trained models as can be seen in table \ref{tab1}. 
A qualitative analysis between methods can be seen in supplementary figure 1, where the three retinal experts (B.A., J.S. and M.H.) annotated all 14 classes on images from the two domains Topcon and Cirrus, as well as on the dataset of the manufacturer Bioptigen. There, accurate segmentations of the SVDNA model can be seen, whereas other methods often struggle to segment correctly.

\section{Discussions and Limitations}
We demonstrate that the minimal SVDNA method outperforms or performs on par with state-of-the-art UDA methods and allows for accurate cross-device segmentation of OCT Images without using any additional labeled data. The other main benefit of the SVDNA method is that it integrates directly to the regular training pipeline of semantic segmentation networks and does not need any modifications to the models architecture or training of a separate style transfer model as done in the CycleGAN approach \cite{Romo-Bucheli2020-az}, which can influence the feasibility of applying a method in practice. One possible limitation of the SVDNA method could be that it does not necessarily denoise images well. This would mean that domain adaptation from Spectralis, a less noisy domain, to Topcon, Cirrus, or Bioptigen works well but the opposite direction might not be as successful. One possible solution would then be to do test time SVDNA style transfer, meaning that when predicting low noise images on a high noise image trained model, one could add noise to the images before feeding it into the model, similar to the idea used by the CycleGAN approach where they restyle images from the target to the source domain at test time.  
The other benchmarked state-of-the-art domain adaptation methods did not consistently perform well for all devices. In our experiments with the CycleGAN model, we sometimes found it can slightly alter the content of the image or sometimes even fail completely to produce images similar to the input OCT image. As OCT biomarkers such as intraretinal cysts and PEDs are often represented by ambiguous and hard to detect textual features, even the slightest distortion to the morphology of the tissue can cause incorrect segmentation results, for examples of content distortions achieved by an optimized CycleGAN see supplementary figure 4. It is worth noting that in other domains such as on natural images, small context distortions might have a smaller effect on the segmentation performance. The features distinguish objects such as cars, houses, and trees or are considerably different from those representing OCT biomarkers. The  FDA method \cite{Yang2020-vd} does not directly distort the content of the source image, however, we were not able to improve meaningfully over the baseline. Depending on the hyperparameter settings, either little to no style transfer was achieved for small beta values or image distorting artifacts got introduced into images for higher beta values, as can be seen in supplementary figure 3. Finally using a self-training-based domain adaptation method did not work well in our experiments which might be due to the limited size of the datasets.
\section*{Acknowledgements}
The present contribution is supported by the Helmholtz Association under the joint research school “Munich School for Data Science - MUDS“. H.S. and F.J.T. acknowledge support by the BMBF (grant number: 031L0210A) and by the Helmholtz Association’s Initiative and Networking Fund through Helmholtz AI (grant number: ZT-I-PF-5-01). We want to thank Dr. Carsten Marr for his support and Dr. Hrvoje Bogunović for his help in evaluating our results on the non-public testset.
\newpage
\printbibliography
\end{document}